\documentclass[11pt]{article}
\usepackage{coling2020}
\usepackage{times}
\usepackage{url}
\usepackage{soul}
\usepackage{latexsym}
\usepackage{graphicx,scalerel}
\usepackage{caption}
\usepackage{booktabs}
\usepackage{tabularx}
\usepackage{makecell}
\usepackage{multirow}
\usepackage{enumitem}

\renewcommand\theadfont{\bfseries}

\renewcommand\theadfont{\bfseries}
\usepackage{etoolbox}
\usepackage{amsthm,amsmath,amsfonts,bm,xspace}
\usepackage{bbm}
\usepackage{color}

\newcommand{\comment}[1]{}

\def\eqref#1{(\ref{#1})}
\def\1{\bm{1}}

\def\valpha{{\bm{\alpha}}}

\def\vb{{\bm{b}}}

\def\vg{{\bm{g}}}

\def\vx{{\bm{x}}}
\def\vy{{\bm{y}}}

\DeclareMathAlphabet{\mathsfit}{\encodingdefault}{\sfdefault}{m}{sl}
\SetMathAlphabet{\mathsfit}{bold}{\encodingdefault}{\sfdefault}{bx}{n}

\newcommand{\softmax}{\mathrm{softmax}}

\usepackage[ruled,vlined]{algorithm2e}
\SetKwRepeat{Do}{do}{while}
\SetAlgoLined
\SetKwInOut{Input}{Input}
\SetKwInOut{Output}{Output}
\usepackage{amsmath,tabstackengine}
\usepackage{eqparbox}
\usepackage{booktabs}
\usepackage{caption}
\usepackage{algpseudocode}
\newlist{myitemize}{itemize}{3}
\usepackage{xcolor}
\usepackage{amsmath}
\usepackage{pgfplots}
\usepackage{pgfplotstable}
\usepackage{subfig}
\usepackage{graphicx}       
\usepackage{tikz}
\usetikzlibrary{matrix,spy,backgrounds}
\usepgfplotslibrary{units}
\definecolor{green}{rgb}{0.0, 0.5, 0.0}

\colingfinalcopy

\title{Collective Wisdom: Improving Low-resource Neural Machine Translation using Adaptive Knowledge Distillation}

\author{
Fahimeh Saleh\\
Monash University\\
{\tt \small first.last@monash.edu}
\And
Wray Buntine\\
Monash University\\
{\tt \small first.last@monash.edu}
\And
Gholamreza Haffari\\
Monash University\\
{\tt \small first.last@monash.edu}
}

\date{}

\newcommand{\reza}[1]{\textcolor{black}{#1}}

\begin{document}
\maketitle
\begin{abstract}
\reza{
Scarcity of parallel sentence-pairs poses a significant hurdle for training high-quality Neural Machine Translation (NMT) models in bilingually low-resource scenarios. 
A standard approach is transfer learning, which involves taking a model trained on a high-resource language-pair and fine-tuning it on the data of the low-resource MT condition of interest. 
However, it is not clear generally which high-resource language-pair offers the best transfer learning for the target MT setting. Furthermore, different transferred models may have complementary semantic and/or syntactic strengths, hence using only one model may be sub-optimal.     
In this paper, we tackle this problem using knowledge distillation, where we propose to distill the knowledge of \emph{ensemble of teacher} models to a single \emph{student} model. 
As the quality of these teacher models varies, we propose an effective adaptive knowledge distillation approach to dynamically adjust the contribution of the teacher models during the distillation process. 
Experiments on transferring from a collection of six language pairs from IWSLT to five low-resource language-pairs from TED Talks demonstrate the effectiveness of our approach, achieving up to +0.9 BLEU score improvement compared to strong baselines. 
}
%In this paper, we propose a two-phase method  to tackle this challenge. The first phase involves transfer learning, where models trained on high-resource languages-pairs are fine-tuned on the data of the low-resource MT condition of interest. 
%
%The second phase involves disstilling the knowledge from this collection of \emph{teachers} to a single \emph{student} model. 
%
%As the quality of these teacher models vary, we propose an adaptive knowledge distillation approach to adaptively adjust the contribution of the teacher models during the training process of the student. 
%
%NMT models, where a pretrained modeld on high-resource data is fine tuned on .  The transferred models are treated as \emph{teachers} which produce soft targets for each low-resource language. In the second phase, we adaptively distil knowledge from all teachers based on their capability to improve the accuracy of the low-resource NMT model (\emph{student}). By optimizing the student to fit the teachers' distribution over smoothed labels, we expect the student’s generalisation affected by teachers' probability calibration. Moreover, we propose to control the teachers' contributions when computing the soft targets for knowledge distillation, such that better teachers contribute more. This contribution is adaptively changing based on how good a teacher captures the context of an incoming mini-batches during training. Experiments on IWSLT and TED dataset demonstrate the effectiveness of our model which outperforms strong baselines on the translation of five low-resource languages to English.

\end{abstract}

\section{Introduction}
\blfootnote{
    %
    % for review submission
    %
    %\hspace{-0.65cm}  % space normally used by the marker
    %Place licence statement here for the camera-ready version. See
    %Section~\ref{licence} of the instructions for preparing a
    %manuscript.
    %
    % % final paper: en-uk version 
    %
    % \hspace{-0.65cm}  % space normally used by the marker
     This work is licensed under a Creative Commons 
     Attribution 4.0 International Licence.
     Licence details:
     \url{http://creativecommons.org/licenses/by/4.0/}.
    % 
    % % final paper: en-us version 
    %
    % \hspace{-0.65cm}  % space normally used by the marker
    % This work is licensed under a Creative Commons 
    % Attribution 4.0 International License.
    % License details:
    % \url{http://creativecommons.org/licenses/by/4.0/}.
}

\reza{
Neural models have been revolutionising machine translation (MT), and have achieved state-of-the-art for many high-resource language pairs \cite{chen2018best,stahlberg2019neural,maruf2019survey}. However, the scarcity of bilingual parallel corpora is still a major challenge for training high-quality NMT models 
% especially for a broad range of languages for which the available translation training resources are too small to be used with existing NMT systems 
\cite{koehn2017six}.
Transfer learning by fine-tuning, from a model trained for a high-resource language-pair,
% \wray{Wray: Having trouble with this:  isn't "high-resource language-pair" mean \textbf{both} source and target are high resource so how does this relate to what we do?}
is a standard approach to tackle the scarcity of the data in the target low-resource language-pair \cite{dabre2017empirical,kocmi2018trivial,saleh2019naver,kim2019effective}.
However, this is a one-to-one approach, which is not able to exploit models trained for multiple high-resource language-pairs for the target language-pair of interest. 
Furthermore, models transferred from different high-resource language-pairs may have complementary syntactic and/or semantic strengths, hence using a single model may be sub-optimal. 
} 

%Transfer learning is one of the widely used solutions for addressing the data scarcity problem in low-resource scenarios \cite{dabre2017empirical,kocmi2018trivial,saleh2019naver,kim2019effective}. 
% However, applying the original transfer learning to LR models is neither able to make full use of highly related multiple high-resource languages nor to receive different parameters from all effective high-resource NMT models simultaneously. 
%However, transfer learning from high-resource to low-resource NMT models is generally a one-to-many approach which is not able to exploit multiple high-resource languages and high-resource NMT models' parameters simultaneously. Contrariwise, 

\reza{
Another appealing approach is multilingual NMT, whereby a single NMT model is trained 
by combining data from multiple high-resource and low-resource language-pairs
\cite{johnson2017google,ha2016toward,neubig2018rapid}.
%
%is an appealing approach for low-resource languages by utilizing the training examples of multiple languages \cite{johnson2017google,ha2016toward,neubig2018rapid}. 
%In practice, for training a multilingual NMT, a multilingual vocabulary set from all language pairs are used for training a single NMT model among all languages that enable sharing resources between high-resource and low-resource languages.
% and improves the regularization of the model by avoiding over-fitting to the limited data of the low-resource languages. 
However, the performance of a multilingual NMT model is highly dependent on the types of languages used to train the model.  Indeed, if languages are from very distant language families, they lead to negative transfer, causing low translation quality in the multilingual system compared to the counterparts trained on the individual language-pairs  \cite{tan2019multilingual1,oncevay2020bridging}. 
To address this problem, \cite{tan2019multilingual2} has proposed a knowledge distillation approach to effectively train a multilingual model, 
by selectively distilling the knowledge from individual teacher models to the multilingual student model. However, still all the language pairs are trained
in a single model with a blind contribution during training.
%during the training process when the accuracy of the individual models surpasses the multilingual one. 
% by distilling knowledge from individual NMT models. To avoid distilling knowledge from the not effective teachers, they selectively apply distillation during the training process when the accuracy of the individual models surpasses the multilingual one. 
}

\reza{
In this paper, we propose a many-to-one transfer learning approach which can effectively transfer models from multiple high-resource language-pairs to a target low-resource language-pair of interest. 
As the fine-tuned models from different high-resource language pairs can have complementary syntactic and/or semantic strengths in the target language-pair, our idea is to distill their knowledge into a single student model to make the best use of these teacher models. 
We further propose an effective adaptive knowledge distillation (AKD) approach to dynamically adjust the contribution of the teacher models during the distillation process, enabling making the best use of teachers in the ensemble. 
Each teacher model provides dense supervision to the student via dark knowledge \cite{darkHinton15} using a mechanism similar to label smoothing \cite{DBLP:conf/cvpr/SzegedyVISW16,DBLP:conf/nips/MullerKH19}, where the amount of smoothing is regulated by the teacher. 
In our AKD approach, the label smoothing coming from different teachers is  combined and regulated, based on the loss incurred by the teacher models during the distillation process.  
%
%\wray{Wray:  This next sentence could be deleted if you need space.}
%Although we focus on the application of this method for NMT, it can be applied more generally to other NLP tasks suffering from the scarcity of training data, e.g. summarisation {CITE}  and question answering \todo{CITE}.
}
%Experimental results on various teacher-student language pairs show up to 0.9 BLEU score improvement compare to the strong baselines.
Experiments on transferring from a collection of six language pairs from IWSLT to five low-resource language-pairs from TED Talks demonstrate the effectiveness of our approach, achieving up to +0.9 BLEU score improvements compared to strong baselines. 

\begin{figure*}
\scriptsize
\captionsetup{font=small}
\begin{tabular}{cc}
\begin{minipage}[t]{0.5\linewidth}\vspace{0pt}
\begin{algorithm}[H]
\captionsetup{font=small}
\Input{$\mathcal{D}_{LR}:=\{(\vx_1,\vy_1),..,(\vx_n,\vy_n)\}$, low resource dataset, Individual models $\{\theta^l\}_{l=1}^{L}$ for $L$ language pairs, Total training epochs: $N$}
\Output{$\theta_{LR}$: low-resource model\\}
Randomly initialize low-resource model $\theta_{LR}$ \; 
%set $\alpha_{l}=1/L \ \ , \ \ \forall l \in [1,..,L]$  \; 
$n = 0$ \;
%i = 0$ \;
\While{$n < N$}{
    $D_{LR} = random\_permute(\mathcal{D}_{LR})$ \;
    $\vb_1,..,\vb_M = create\_minibatches(\mathcal{D}_{LR})$ \;
    $m = 1$ \;
    \While{$m \le M$}{
    // compute contribution weights\;
      \For{$l \in L$}{
        %   $\Delta_{l} = acc(\theta_{LR})- acc(\theta^l)$ \;
        $\Delta_{l} = - ppl(\theta^l(b_m))$ \;
          }
    % $\valpha = (1-\beta) \valpha + \beta \softmax (\Delta_{1},..,\Delta_L)$ \; 
     $\valpha = \softmax (\Delta_{1},..,\Delta_L)$ \; 
     // compute the gradient \;
      $\vg = \nabla_{\theta_{LR}} \mathcal{L}_{ALL}^{adaptive}(\vb_m,\theta_{LR},\{\theta^l\}_{1}^L,\valpha)$  \;
      // updates the parameters using the optimiser ADAM \;
      $\theta_{LR} = \textrm{update\_param}(\theta_{LR},\vg) $ \; 
     $m = m + 1$ \; 
     }
    $n = n + 1$ \;
}
\caption{}
\label{ref:algorith}
\end{algorithm}
\end{minipage}
\hspace{0.4cm}
\begin{minipage}[t]{0.45\linewidth}\vspace{0pt}
\centering
\includegraphics[width=\textwidth]{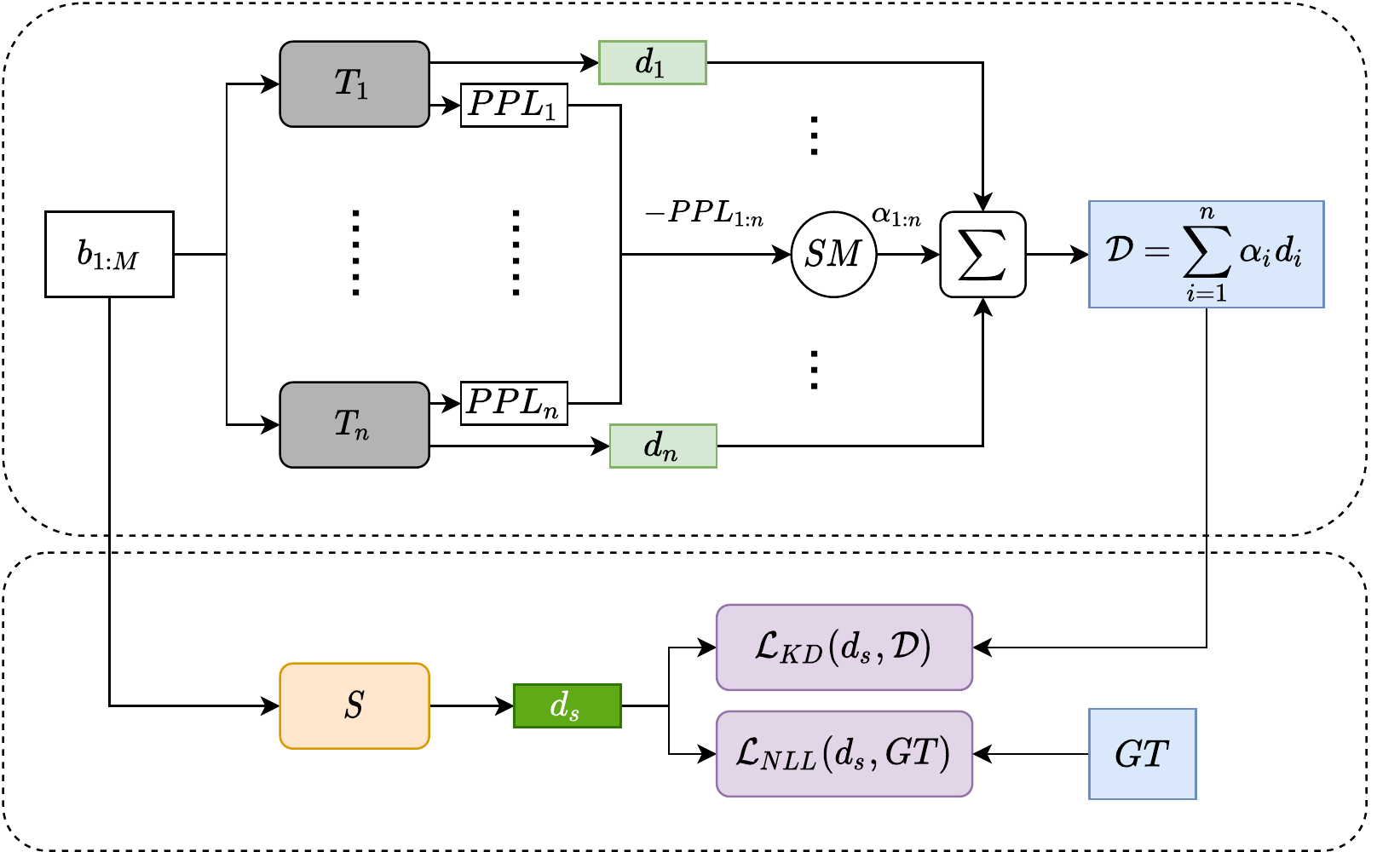}
\caption{Adaptive Knowledge Distillation. \textbf{(Top)} Teachers' contribution weight calculation. $T_{1:n}$ and $d_{1:n}$ denote the freezed teacher models and their corresponding probability distributions respectively. \textbf{(Bottom)} Training the student with adaptive knowledge distillation. $S$,  $SM$, and $GT$ denote the student model, softmax function, and ground-truth respectively.}
\label{fig:method}
\end{minipage}
\vspace{-10pt}
\end{tabular}
\end{figure*}
% \section{Low-Resource NMT with Adaptive Knowledge Distillation}
\section{Adaptive Knowledge Distillation}
\label{approach}
% \subsection{Problem Formulation and the Approach}
We address the problem of low-resource NMT, assuming that we have access to models for high resource languages, and data for low resource model. 
Our approach relies on two main steps, (i) Transferring from high-resource to low-resource language-pairs by fine tuning the high-resource models using the small amount of bilingual data, and (ii) Adaptive distillation of knowledge from the teacher models to the student model. 

More specifically, given a training dataset for a low-resource language-pair, $\mathcal{D}_{LR}:=\{(\vx_1,\vy_1),..,(\vx_n,\vy_n)\}$ and multiple individual high resource NMT models $\{\theta^l\}_{l=1}^{L}$ fine-tuned on $\mathcal{D}_{LR}$ (teachers), we are interested in training a single NMT model (student) by adaptively distilling knowledge from all teachers based on their effectiveness to improve the accuracy of the student. 
Knowledge distillation (KD) is a process of improving the performance of a simple \textit{student} model by using a distribution over soft labels obtained from an expert \textit{teacher} model instead of hard ground-truth labels \cite{darkHinton15}. The training objective to distill the knowledge from a single teacher  to the student involves,
\begin{align}
{\setstackgap{L}{21pt}\ensurestackMath{\tabbedCenterstack[c]{
     -\sum_{\vx,\vy \in \mathcal{D}_{LR}}  \sum_{t=1}^{|\vy|} \sum_{v \in V} Q(v | \vy_{<t} , \vx , \theta^l) \log P(v | \vy_{<t} , \vx , \theta_{LR}) 
    }}}
\end{align}
where $\theta^l$ and $\theta_{LR}$ are the parameters of the teacher and student models, respectively. $P(.\mid .)$ is the conditional probability with the student model and $Q(.\mid .)$ denotes the output distribution of the teacher model. 
%
% This provides dense training signal as \emph{each} word in the vocabulary ($V$) contributes to the training objective, regulated by a weight coming from the teacher. 
According to Equation 1, knowledge distillation provides dense training signal as \emph{each} word in the vocabulary ($V$) contributes to the training objective, regulated by a weight coming from the teacher.
This is in contrast to the negative log-likelihood training objective, which only provides supervision signal based on the correct target words according to the bilingual training data,
\vspace{-10pt}
\begin{align}
{\setstackgap{L}{21pt}\ensurestackMath{\tabbedCenterstack[c]{
    \mathcal{L}_{NLL} (\mathcal{D}_{LR}, \theta_{LR}) := -\sum_{\vx,\vy \in \mathcal{D}_{LR}} \sum_{t=1}^{|\vy|} \log P(y_t | \vy_{<t} , \vx , \theta_{LR}).}}}
\end{align}
Given a collection of teacher models $\{\theta_l\}_{l=1}^L$, we pose the following training objective, 
\vspace{-4pt}
\begin{align}
{\setstackgap{L}{21pt}\ensurestackMath{\tabbedCenterstack[c]{
    \mathcal{L}_{KD}^{adaptive} (\mathcal{D}_{LR}, \theta_{LR},  \{\theta^l\}_{1}^{L},\valpha) := \\ -\sum_{\vx,\vy \in \mathcal{D}_{LR}} \sum_{l=1 }^{L} \alpha_{l} \sum_{t=1}^{|\vy|} \sum_{v \in V} Q(v | \vy_{<t} , \vx , \theta^l) \log P(v | \vy_{<t} , \vx , \theta_{LR}) 
    }}}&&
\end{align}
where $\alpha_l$ regulates the contribution of the $l$-th teacher.
We dynamically adjust the contribution weights over the course of the distillation process, in order to effectively address the knowledge gap of the student during the training process. 
This is achieved based on the rewards (negative perplexity) attained by the teachers on the data, where these values are passed through a softmax transformation to turn into a distribution.  
To stabilize these contribution weights over the course of the training process, we smooth them using a running geometric average. 
\reza{
The student model is trained end-to-end with a weighted combination of losses coming from the ensemble of teachers and the data,
\begin{align}
      \mathcal{L}_{ALL}^{adaptive} (\mathcal{D}_{LR}, \theta_{LR},  \{\theta^l\}_{1}^{L}, \valpha) &:=& {\lambda}_1 \mathcal{L}_{NLL} (\mathcal{D}_{LR}, \theta_{LR}) +  {\lambda}_2 \mathcal{L}_{KD}^{adaptive} (\mathcal{D}_{LR}, \theta_{LR},  \{\theta^l\}_{1}^{L},\valpha) 
\end{align}
where ${\lambda}_1 = 0.5$  and  ${\lambda}_2$ is started from 0.5 and gradually increased to 3 following the annealing function of~\cite{bowman2015generating} in our experiments.  Our approach is summarized in Algorithm \ref{ref:algorith} and Figure~\ref{fig:method}.}

\section{Experiments}
\subsection{Settings}
% \normalsize
\paragraph{Data.} We conduct our experiments on the European languages of IWSLT and TED datasets. The language pairs with more than 100K training data are considered as high-resource and the ones less than 15k are assumed as low-resource. The high-resource models are trained on IWSLT2014 (ru,de,it,pl,nl,es-en).  
IWSLT 2014 MT task data (sl-en) \cite{cettolo2014report}, and TED talk data (gl/et/nb/eu-en)  \cite{qi2018and} are used as low-resource languages. Detail about the preprocessing step and the statistics of data and language codes based on ISO 639-1 standard\footnote{http://www.loc.gov/standards/iso639-2/php/English\_list.php} are listed in Section 1.1 of Appendix A. 

% %%%%%%%%%%%%%%%%%
% \input{table1-data}
% %%%%%%%%%%%%%%%%%

\normalsize
\paragraph{Training configuration.} Individual low-resource and high-resource NMT models are trained on the low-resource data. The first trained from scratch and the later by finetuning with the vanilla transformer architecture. For multilingual NMT, we train a single model with all high-resource and the up-sampled of low-resource language pairs by using a decoder language embedding layer to identify the type of language during the inference step. Multilingual selective knowledge distillation \cite{tan2019multilingual2} is trained with all language pairs while matching the outputs of each low-resource model simultaneously through knowledge distillation. For training our approach, we fine-tune the high-resource models with low-resource languages and treat them as teachers. When training on the low-resource language, we load teacher models into memory and train a single low-resource model (student) from scratch while using the weighted average of teachers' probabilities based on their contribution weight. In order to make clear how different teachers contribute during training the student, we illustrate contribution weights of all teachers for first 30 iterations of different mini-batches during the training in Figure 2.

\paragraph{Model configuration.}
All models are trained with Transformer architecture \cite{vaswani2017attention}, with the model hidden size of 256, feed-forward hidden size of 1024, and 2 layers,  implemented in Fairseq framework \cite{ott2019fairseq}. 
We use the Adam optimizer \cite{kingma2014adam}  and an inverse square root schedule with warm-up (maximum LR 0.0005). We apply dropout and label smoothing with a rate of 0.3 and 0.1 respectively. The source and target embeddings are shared and tied with the last layer. We train with half-precision floats on one V100 GPU, with at most 4028 tokens per batch. 
% and delayed updates of 10 batches. 
% When fine-tuning on low-resource language we use a fixed learning rate schedule (Adam with 0.00005 LR) and the same batch size as before on a single GPU without delayed updates.

% All models are trained with Transformer architecture \cite{vaswani2017attention}, with the model hidden size of 256, feed-forward hidden size of 1024, and 2 layers,  implemented in Fairseq framework \cite{ott2019fairseq}. 
% We use the Adam optimizer \cite{kingma2014adam}  and an inverse square root schedule with warmup (maximum LR 0.0005). We apply dropout and label smoothing with a rate of 0.3 and 0.1. The source and target embeddings are shared and tied with the last layer. We train with half-precision floats on one V100 GPU, with at most 4028 tokens per batch.

%%%%%%%%%%%%%
% \input{table2-results}
%%%%%%%%%%%%%

\subsection{Results}
In Table \ref{tab:comparison}, we compare our approach with individual NMT models, transferred models from high-resource language pairs, multilingual NMT, and multilingual selective knowledge distillation \cite{tan2019multilingual2}. We selected the best models according to the SacreBLEU\footnote{SacreBLEU signature: BLEU+case.mixed+numrefs.1+
smooth.exp+tok.none+version.1.3.1} score on the validation set.
In our experiments, bold numbers indicate the best results and underlined numbers show the second best ones.
Transfer learning results are inline with the language family relationships \cite{littell2017uriel}. The high-resource languages which are linguistically close to the low-resource languages have the most impact on low-resource model's improvement. Likewise, the contribution weights of different teachers are consistent with the performance of the teachers as hypothesized (See results in Table \ref{tab:comparison} and Figure 2). 
According to Table \ref{tab:comparison}, the multilingual models (with and without knowledge distillation) are less accurate than at least one of the transferred models from high-resource languages\footnote{Except for the Basque language which is extremely low-resource and is linguistically as distant to all the languages in the multilingual setting.}. This suggests a weak link may exist between the impact of each high-resource language and its contribution during the training multilingually. Adaptive knowledge distillation compensates this blind collaboration between teachers by weighting the teachers' contributions particularly for the cases where majority of teachers and student are linguistically close such as ``nb-en". The qualitative examples are presented in Section 1.4 of Appendix A. 
It is worth noting that, we empirically observed when there is more diversity in teachers (e.g, in case of ``gl-en" in Table~\ref{tab:comparison}), adaptive KD underperforms compared to the best teacher and we hypothesise this happens because there is an empirically dominant teacher (``es"). 
%It is worth noting that, we empirically observed when there is more diversity in teachers (e.g., one strong teacher while other teachers are quite weak), adaptive KD underperforms compared to the best teacher (e.g, ``gl-en") and we hypothesise this happens because there is an empirically dominant teacher, "es". For instance, in case of  in Table \ref{tab:comparison}.)
This observation suggests that a prior effort for choosing the proper teacher languages (e.g., based on the language family information) will directly impact the performance of the low-resource NMT model.

\section{Analysis}
\label{sec:analysis}
% \paragraph{Contribution weight analysis.}
% In order to make clear how different teachers contribute to train the student (low-resource language), we illustrate the contribution weights of all teachers for first 30 iterations of different mini-batches while training the low-resource NMT model (Figure \ref{fig:contrib-wights}). 

% \input{fig2-weights}

%%%%%%%%%%%%%%%%%%

\renewcommand\theadfont{}
%\begin{table}[t]
%    \centering
%%    \captionsetup{font=small}
%\scalebox{0.8}{
% \begin{tabular}{|c|c|c|c|c|c|c|c|c|c|c|}
%    \hline 
%    \multirow{1}{*}{pairs}&Individual&\multicolumn{6}{c|}{Transfer Learning from (x-en) model}&Multilingual &\thead{Multilingual Selective KD\\\cite{tan2019multilingual2}} &Adaptive KD\\ 
%        \cline{3-8}
%         &  & ru & de & it & es & pl & nl &  &  & \\
%         \hline
%         sl-en & 10.58 & 10.36 & 14.09 & 13.29 & 16.89 & \textbf{17.63} & 16.67 & 15.97 & % & \textbf{18.05} \textbf{}\\ 
%        \hline
%    nb-en &26.38 & 20.72 & \textbf{29.67} & 27.24&27.02 &25.00 & 29.21&  & & \textbf{30.41} \\ 
%        \hline 
%       gl-en & 13.87& 11.88 & 17.66 &21.90 &\textbf{27.49} &16.67 & 17.05& 25.27 & & \\ 
%        \hline 
%         eu-en &6.50 & 7.57 & 8.15 &7.93 &  \textbf{9.6}&8.35 & 7.36 & 10.11 &  & \textbf{10.52}\\ 
%        \hline 
%         be-en &4.04 &\textbf{15.10}& 6.66 & 6.45&7.28 &5.80 &5.75 & 13.78 & & \\ 
%        \hline 
%    \end{tabular} 
%    }
%    \caption{BLEU scores of 5 languages$\rightarrow$ English}
%    \label{tab:comparison}
%\end{table}

\begin{table}[t]
    \centering
    \captionsetup{font=small}
\scalebox{0.85}{
 \begin{tabular}{l||c|cccccc|cc|c}
    \toprule%\hline 
    \multirow{1}{*}{\scriptsize{MT Task}}& Individual &\multicolumn{6}{c|}{Transfer Learning from (x$\rightarrow$en) model}&\multicolumn{2}{c|}{Multi-Lingual}  & Multi-Teacher\\ 
        %\cline{3-8}
         x$\rightarrow$en &  student & ru & de & it & es & pl & nl &  Uniform & Selec. KD & Adap. KD\\
         \midrule %\hline \hline 
         sl & 10.58 & 10.36 & 14.09 & 13.29 & 16.89 & \underline{17.63} & 16.67 & 15.97 &  16.17& \textbf{18.35} \textbf{}\\ 
     %   \hline
    nb &26.38 & 32.24 & 32.77 & 31.90 & 30.04 &30.66 & \underline{ 32.86} & 30.06 & 31.08& \textbf{33.72} \\ 
    %    \hline 
       gl & 13.87& 11.88 & 17.66 &21.90 &\textbf{27.49} &16.67 & 17.05& \underline{25.27} & 25.08& 24.50\\ 
    %    \hline 
         eu &6.50 & 9.54 & 10.68 &9.92 &  11.00&10.50 & 10.02 & 10.11 & \underline{11.03} & \textbf{11.38} \\ 
     %   \hline 
         et &10.15 &12.18& 14.85 & 14.93& \underline{15.53} & 14.25& 13.66& 14.91 & 15.15& \textbf{16.20}\\ 
     %   \hline
    %   fi &8.76&11.46 &10.02& \textbf{13.39} & 13.33& 9.85 & 13.25&  & \\
     \bottomrule
    \end{tabular} 
    }
     \vspace{-.2cm}
    \caption{BLEU scores of the translation tasks from five languages into English. Selective KD is based on \cite{tan2019multilingual2}.}
    \label{tab:comparison}
\end{table}
%%%%%%%%%%%%%%%%%%%

\begin{table}[!t]
\vspace{-0.2cm}
\captionsetup{font=small}
    \centering
    \captionsetup{justification=raggedright}
    \begin{minipage}[b!]{0.49\textwidth}
    \centering
    \hspace{0.6cm}
    \scalebox{0.77}{
    \begin{tabular}{|l | l l|}
        \toprule
        Contribution weight setting & gl-en & nb-en\\
        \midrule
%        soft adaptive & 23.89& 33.72  \\
%        hard adaptive & 24.50 & 33.50\\
%        equal contribution & 19.10& 32.60\\
         Adaptive contribution & 24.50& 33.72  \\
%        hard adaptive & 24.50 & 33.50\\
        Equal contribution & 19.10& 32.60\\
        \bottomrule
    \end{tabular}}
    \vspace{-.1cm}
    \caption{Effect of different contribution settings.}
    \label{tab:Anls_contrib_weight}
    \end{minipage}
   \hspace{-0.7cm}
    \begin{minipage}[b!]{0.49\textwidth}
    \centering
    \scalebox{0.77}{
    \begin{tabular}{|l | l l|}
        \toprule
        Contribution temperature  & eu-en  & sl-en\\
        \midrule
         with adaptive temp & 11.38 & 18.35\\ 
         without temp &10.52 & 18.05\\ 
        \bottomrule
    \end{tabular}}
    \vspace{-.1cm}
    \caption{Effect of adaptive temperature.}
    \label{tab:Anls_temp}
    \end{minipage}
    \vspace{-12pt}
\end{table}
%%%%%%%%%%%%%%%%%%
%  \noindent \textbf{Contribution weight settings.}
 \subsection{Contribution Weight Analysis}
 To analyse the effect of teachers' contribution weights, we compare two different contribution settings:
 \emph{(i) Adaptive contribution:} which assigns the contribution weights to all the teachers based on their performance per mini-batch as explained in Section \ref{approach}. \emph{(ii) Equal contribution:} which gives all the teachers the same contribution weights. According to Table \ref{tab:Anls_contrib_weight}, the equal contribution setting is not as effective as the adaptive contribution especially for the languages with more inconsistent  teachers (based on BLEU score) e.g., ``gl-en".
 
%  To analyse the effect of teachers' contribution weights, we compare two contribution policies:
%  \emph{a) Soft adaptive:} which assigns the contribution weights to all the teachers based on their performance per mini-batch as explained in Section \ref{sec:method}.  \emph{b) Hard adaptive:} which chooses one teacher by multinomial sampling from the $\alpha$ distribution per mini-batch. \emph{c) Equal contribution:} which gives all the teachers the same contribution weights. 

% According to Table \ref{tab:comparison} and \ref{tab:Anls_contrib_weight}, the worst contribution weight setting is for \textit{equal contribution} especially for the languages with more inconsistent  teachers (based on BLEU score) e.g., ``gl-en" where \textit{hard adaptive} performs better. However, in other cases with more consistent teachers (based on BLEU score), \textit{soft adaptive} performs the best (e.g.,``nb-en").

% \noindent \textbf{Contribution temperature scaling.} 
\subsection{Contribution Temperature Scaling}
Through the experiments, we observed that when most of the teachers do not agree (in terms of perplexity),
% on the context of a mini-batch, 
a constant temperature is not an ideal option.
% ; weighing a \textit{bad} teachers similar to a \textit{good} one.
An alternative is to adaptively change the value of the temperature given the \textit{agreement} among the teachers determined based on the distance between the maximum and minimum perplexity between teachers which can be formulated as:
%as $ \tau=\frac{1 - (\text{max}(S) - \text{min}(S))}{N}  \label{eq:contrib_t}$ 
\begin{align}
    \tau=\frac{1 - (\text{max}(S) - \text{min}(S))}{N} 
    \label{eq:contrib_t}
\end{align}
where $S$ is the output of the softmax operation on the negative perplexity of all $N$ teachers and ($\text{max}$(S)$ - \text{min}$(S)$)$ is inversely proportional to the extent of the agreement between teachers. Such temperature scaling encourages the contribution of better teachers in case of the existence of a disagreement, while it allows similar contributions when all teachers agree on a mini-batch. Table \ref{tab:Anls_temp} shows the effect of adaptive temperature for two languages.
%%%%%%%%%%%%%%%%%
%\input{tables34-analysis}
%%%%%%%%%%%%%%%%%
% \paragraph{Contribution weight analysis.}
% \label{sec:app-contrib-weight-analysis} We illustrate the contribution weights of all teachers for the first 30 iterations of different mini-batches while training the low-resource NMT model in Figure \ref{fig:contrib-wights}. This visualisation shows how different teachers contribute to train the student. As it is also shown earlier in the results, for those cases where most of the teachers are linguistically close like ``sl-en" and ``nb-en", teachers' contribution weights are very close together and our approach outperforms other baselines by adaptively assigning the contribution weights to the teachers during the knowledge distillation process.
\pgfplotstableread{
itr	ru	de	it	es	pl	nl
1	0.0007	0.2511	0.0503	0.5488	0.0819	0.0672
2	0.0121	0.2189	0.1053	0.412	0.1421	0.1096
3	0	0.096	0.0243	0.8065	0.0471	0.0261
4	0.0001	0.1671	0.0152	0.7663	0.0382	0.0132
5	0	0.1507	0.0185	0.7805	0.0289	0.0213
6	0.0002	0.1231	0.0234	0.7922	0.0373	0.0238
7	0	0.0734	0.006	0.8921	0.0191	0.0094
8	0.0001	0.1444	0.0284	0.7686	0.0315	0.0271
9	0.0002	0.1709	0.023	0.7286	0.0502	0.027
10	0	0.0874	0.0169	0.8551	0.0345	0.0061
11	0.0003	0.1092	0.0184	0.8238	0.0214	0.0269
12	0	0.2783	0.007	0.639	0.0493	0.0264
13	0.0001	0.1539	0.0147	0.7886	0.0246	0.0182
14	0.0001	0.2036	0.0279	0.7013	0.0377	0.0293
15	0	0.1907	0.0173	0.7467	0.0301	0.0152
16	0.0001	0.1954	0.0281	0.715	0.0314	0.0299
17	0	0.1618	0.0215	0.7474	0.0511	0.0182
18	0	0.0684	0.0111	0.8612	0.0515	0.0078
19	0.0003	0.2154	0.0385	0.6615	0.0482	0.036
20	0.0001	0.1852	0.0429	0.6968	0.0458	0.0291
21	0.0005	0.1562	0.0416	0.6695	0.0952	0.0369
22	0.0001	0.1206	0.0258	0.8207	0.022	0.0108
23	0	0.1749	0.0162	0.7748	0.024	0.0101
24	0.0001	0.199	0.026	0.6797	0.0686	0.0266
25	0.0001	0.1389	0.0112	0.7946	0.0401	0.0151
26	0.0001	0.0426	0.0079	0.9383	0.0035	0.0076
27	0	0.1258	0.0088	0.8281	0.0201	0.0171
28	0.0001	0.0754	0.0181	0.8747	0.0158	0.0161
29	0	0.1374	0.0235	0.7917	0.025	0.0223
30	0	0.1749	0.0162	0.7748	0.024	0.0101
}\dataeu
\pgfplotstableread{
itr	ru	de	it	es	pl	nl
1	0.0358	0.3166	0.1504	0.0903	0.134	0.2729
2	0.0284	0.3084	0.1686	0.0974	0.1394	0.2577
3	0.0194	0.3596	0.1516	0.0857	0.1377	0.2461
4	0.0555	0.2659	0.1713	0.1221	0.1622	0.2229
5	0.0195	0.3702	0.1306	0.0613	0.1255	0.2929
6	0.0192	0.3116	0.1503	0.09	0.1285	0.3003
7	0.0165	0.3536	0.1539	0.0837	0.1313	0.2609
8	0.0079	0.3544	0.1226	0.0558	0.128	0.3312
9	0.0212	0.3003	0.1712	0.0919	0.1309	0.2845
10	0.0161	0.3463	0.1534	0.07	0.1258	0.2884
11	0.0185	0.3382	0.1592	0.0972	0.1381	0.2488
12	0.0136	0.31	0.1672	0.09	0.1298	0.2896
13	0.0114	0.3736	0.1218	0.0628	0.1193	0.311
14	0.0151	0.3371	0.1429	0.0692	0.1534	0.2822
15	0.0291	0.3399	0.1498	0.1019	0.1404	0.2389
16	0.0276	0.3248	0.1587	0.0904	0.1426	0.2559
17	0.0185	0.3425	0.1478	0.0906	0.1341	0.2665
18	0.0254	0.3002	0.1579	0.0908	0.1554	0.2704
19	0.0213	0.3276	0.1807	0.0802	0.1446	0.2455
20	0.0242	0.3638	0.1303	0.0829	0.1325	0.2664
21	0.031	0.3238	0.1604	0.0933	0.1372	0.2543
22	0.1056	0.2128	0.1702	0.1437	0.1663	0.2014
23	0.0538	0.2532	0.1707	0.1225	0.171	0.2288
24	0.0038	0.3493	0.142	0.1052	0.1036	0.296
25	0.0093	0.3867	0.1264	0.0653	0.1235	0.2889
26	0.0493	0.2718	0.1725	0.1201	0.1536	0.2326
27	0.0117	0.381	0.1352	0.0925	0.1196	0.26
28	0.039	0.2688	0.1702	0.1087	0.1546	0.2588
29	0.0166	0.3582	0.15	0.0641	0.1286	0.2826
30	0.0227	0.3364	0.1513	0.0853	0.1302	0.2741
}\datanb
\pgfplotstableread{
itr	ru	de	it	es	pl	nl
1	0.9997	0	0	0.0003	0	0
2	0.9998	0	0	0.0002	0	0
3	0.9989	0	0	0.0011	0	0
4	0.9999	0	0	0.0001	0	0
5	0.7466	0.0477	0.0207	0.1211	0.0339	0.03
6	1	0	0	0	0	0
7	0.9987	0.0001	0	0.0012	0	0
8	1	0	0	0	0	0
9	0.9998	0	0	0.0001	0	0
10	0.9991	0.0001	0	0.0008	0	0
11	0.9999	0	0	0.0001	0	0
12	1	0	0	0	0	0
13	1	0	0	0	0	0
14	1	0	0	0	0	0
15	0.9999	0	0	0.0001	0	0
16	0.9999	0	0	0.0001	0	0
17	1	0	0	0	0	0
18	0.9999	0	0	0.0001	0	0
19	0.9912	0.001	0.0001	0.0073	0.0002	0.0002
20	1	0	0	0	0	0
21	1	0	0	0	0	0
22	0.9999	0	0	0.0001	0	0
23	1	0	0	0	0	0
24	0.9997	0	0	0.0003	0	0
25	1	0	0	0	0	0
26	0.9958	0.0006	0.0001	0.0034	0.0001	0.0001
27	1	0	0	0	0	0
28	0.9986	0.0001	0	0.0013	0	0
29	0.9993	0	0	0.0007	0	0
30	0.9992	0	0	0.0007	0	0
}\databe
\pgfplotstableread{
itr	ru	de	it	es	pl	nl
1	0.0764	0.1434	0.103	0.2177	0.2293	0.2302
2	0.0896	0.1525	0.0957	0.2123	0.226	0.2238
3	0.0796	0.1527	0.1038	0.2126	0.2243	0.227
4	0.0724	0.1501	0.0989	0.2109	0.2283	0.2394
5	0.0779	0.1471	0.0955	0.2146	0.2304	0.2346
6	0.0798	0.1612	0.0952	0.2061	0.2251	0.2326
7	0.1078	0.1563	0.1238	0.2	0.2043	0.2076
8	0.1072	0.1552	0.1113	0.1985	0.2176	0.2101
9	0.0852	0.1437	0.0965	0.2227	0.2227	0.2293
10	0.0961	0.1558	0.1203	0.1979	0.2166	0.2133
11	0.0834	0.1428	0.0939	0.2191	0.2302	0.2306
12	0.0927	0.1466	0.0939	0.2143	0.2248	0.2276
13	0.0791	0.1529	0.0835	0.2207	0.2274	0.2365
14	0.1053	0.156	0.121	0.2018	0.2074	0.2085
15	0.0765	0.1421	0.0893	0.2229	0.2402	0.229
16	0.0865	0.1446	0.0957	0.2166	0.226	0.2306
17	0.0864	0.1505	0.099	0.2111	0.2236	0.2294
18	0.0917	0.1491	0.0979	0.2098	0.2225	0.229
19	0.0923	0.1482	0.0979	0.2085	0.2219	0.2312
20	0.0812	0.1532	0.1071	0.2066	0.2238	0.2281
21	0.0997	0.1564	0.11	0.2118	0.2105	0.2116
22	0.0752	0.144	0.0984	0.2117	0.2295	0.2412
23	0.0784	0.1484	0.0933	0.2144	0.2274	0.2382
24	0.0908	0.1456	0.0895	0.2149	0.2313	0.2279
25	0.0859	0.1416	0.1039	0.2203	0.2257	0.2226
26	0.0661	0.1536	0.0877	0.2212	0.2297	0.2417
27	0.0904	0.152	0.0827	0.2213	0.2235	0.2301
28	0.0918	0.1532	0.098	0.2126	0.2206	0.2238
29	0.0887	0.1568	0.053	0.2173	0.239	0.2452
}\datasl
\pgfplotstableread{
itr	ru	de	it	es	pl	nl
1	0.0002	0.0296	0.1622	0.7727	0.019	0.0163
2	0.054	0.1486	0.2088	0.3321	0.1253	0.1313
3	0.0002	0.0378	0.1324	0.789	0.0194	0.0212
4	0.0001	0.0332	0.1204	0.8199	0.0138	0.0125
5	0.0003	0.0354	0.167	0.7602	0.0181	0.019
6	0.0001	0.0256	0.1449	0.8016	0.0112	0.0165
7	0.0002	0.0385	0.1507	0.776	0.0178	0.0168
8	0.0002	0.0228	0.1366	0.8115	0.0144	0.0145
9	0.0002	0.0232	0.145	0.8051	0.0127	0.0137
10	0.0003	0.0267	0.1278	0.8076	0.0171	0.0205
11	0.0084	0.1027	0.2227	0.5268	0.0677	0.0716
12	0.0006	0.0415	0.1728	0.7261	0.0303	0.0287
13	0.0003	0.0382	0.1702	0.7411	0.0249	0.0253
14	0.0003	0.0239	0.1675	0.7733	0.0165	0.0186
15	0.0018	0.061	0.1986	0.6651	0.0386	0.0348
16	0.0021	0.0652	0.2005	0.6434	0.0424	0.0465
17	0.0009	0.0384	0.2042	0.6886	0.0307	0.0372
18	0.0002	0.0201	0.1231	0.8152	0.019	0.0224
19	0.0006	0.0334	0.1962	0.7212	0.0272	0.0214
20	0	0.0114	0.0976	0.8793	0.0044	0.0073
21	0.0012	0.0459	0.1904	0.6953	0.0352	0.032
22	0	0.0097	0.0587	0.9217	0.0035	0.0065
23	0.0002	0.0296	0.1445	0.7953	0.0167	0.0138
24	0.0001	0.0173	0.1467	0.8135	0.0098	0.0127
25	0.002	0.0665	0.1944	0.6438	0.0465	0.0469
26	0.0015	0.0461	0.1949	0.69	0.035	0.0326
27	0	0.0218	0.0939	0.8615	0.0081	0.0146
28	0.0015	0.0472	0.1493	0.7484	0.0241	0.0295
29	0.0057	0.0781	0.2251	0.566	0.0611	0.0641
30	0.0003	0.0302	0.146	0.7852	0.0192	0.0191
}\datagl

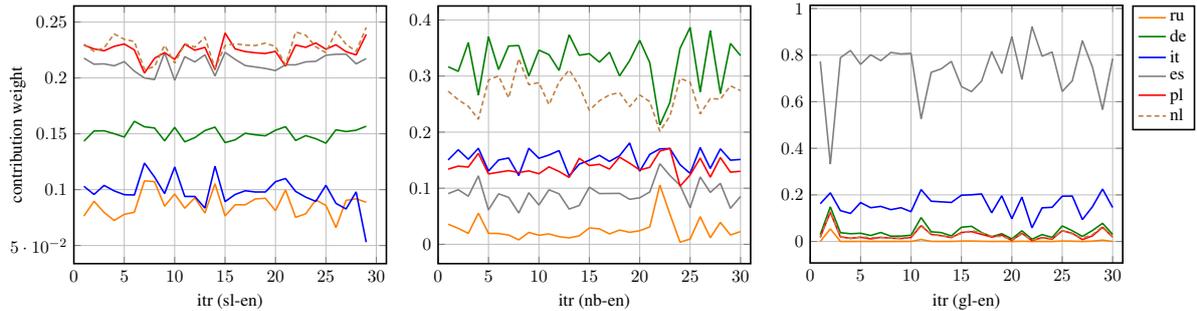
\begin{figure}[t]
\begin{minipage}[b]{0.3\textwidth}
\begin{tikzpicture}[spy using outlines={circle, magnification=2, size=0.4cm, connect spies},scale = 0.6]
\pgfplotsset{
    every axis/.append style={line width=1pt},
    use units = true,
    xlabel = itr (sl-en),
    ylabel = contribution weight,
    xmin=0,
    xmax=31,
    cycle list name=exotic,
}
\begin{axis}[grid=major,cycle list name=exotic,]
    \addplot+[orange,line join=bevel,line width=1pt,mark=none] table[y=ru]{\datasl};
    \addplot+[green,line join=bevel,line width=1pt,mark=none] table[y=de]{\datasl};
    \addplot+[blue,line join=bevel,line width=1pt,mark=none] table[y=it]{\datasl};
    \addplot+[gray,line join=bevel,line width=1pt,mark=none] table[y=es]{\datasl};
    \addplot+[red,line join=bevel,line
    width=1pt,mark=none] table[y=pl]{\datasl};
     \addplot+[brown,line join=bevel,line width=1pt,mark=none] table[y=nl]{\datasl};
\end{axis}
\end{tikzpicture}
\end{minipage}
\captionsetup{font=small}
\begin{minipage}[b]{0.3\textwidth}
\begin{tikzpicture}[spy using outlines={circle, magnification=2, size=0.4cm, connect spies},scale = 0.6]
\pgfplotsset{
    every axis/.append style={line width=1pt},
    use units = true,
    xlabel = itr (nb-en),
    %ylabel = contribution weight,
    xmin=0,
    xmax=31,
    cycle list name=exotic,
}
\begin{axis}[grid=major,cycle list name=exotic,]
    \addplot+[orange,line join=bevel,line width=1pt,mark=none] table[y=ru]{\datanb};
    \addplot+[green,line join=bevel,line width=1pt,mark=none] table[y=de]{\datanb};
    \addplot+[blue,line join=bevel,line width=1pt,mark=none] table[y=it]{\datanb};
    \addplot+[gray,line join=bevel,line width=1pt,mark=none] table[y=es]{\datanb};
    \addplot+[red,line join=bevel,line width=1pt,mark=none] table[y=pl]{\datanb};
     \addplot+[brown,line join=bevel,line width=1pt,mark=none] table[y=nl]{\datanb};
\end{axis}
\end{tikzpicture}
\end{minipage}
\begin{minipage}[b]{0.3\textwidth}
\begin{tikzpicture}[spy using outlines={circle, magnification=2, size=0.4cm, connect spies},scale = 0.6]
\pgfplotsset{
    every axis/.append style={line width=1pt},
    use units = true,
    xlabel = itr (gl-en),
    %ylabel = contribution weight,
    xmin=0,
    xmax=31,
    legend style={
       cells={anchor=west},
        legend pos=outer north east,
    },
    cycle list name=exotic,
    legend entries={ru,de,it,es,pl,nl}
}
\begin{axis}[grid=major,cycle list name=exotic,]
    \addplot+[orange,line join=bevel,line width=1pt,mark=none] table[y=ru]{\datagl};
    \addplot+[green,line join=bevel,line width=1pt,mark=none] table[y=de]{\datagl};
    \addplot+[blue,line join=bevel,line width=1pt,mark=none] table[y=it]{\datagl};
    \addplot+[gray,line join=bevel,line width=1pt,mark=none] table[y=es]{\datagl};
    \addplot+[red,line join=bevel,line
    width=1pt,mark=none] table[y=pl]{\datagl};
     \addplot+[brown,line join=bevel,line width=1pt,mark=none] table[y=nl]{\datagl};
\end{axis}
\end{tikzpicture}
\end{minipage}

\captionsetup{font=small}
\caption{Teachers' contribution weights during the training of low-resource NMT models for ``sl-en", ``gl-en", and ``nb-en" language pairs, first 30 iterations for different mini-batches.}
\label{fig:contrib-wights}
\end{figure}

\section{Conclusion}
In this paper, we present an adaptive knowledge distillation approach to improve NMT for low-resource languages. We address the inefficiency of the original transfer learning and multilingual learning by making wiser use of all high-resource languages and models in an effective collaborative learning manner. Our approach shows its effectiveness in translation of low-resource languages especially when there are complementary knowledge in multiple high-resource languages from the same linguistic family and it is not explicitly clear which language has more impact in every mini-batch of low-resource training data. Experiments on the translation of five extremely low-resource languages to English show improvements compared to the strong baselines.

\section*{Acknowledgements}
This work was supported by
the Multi-modal Australian ScienceS Imaging and
Visualisation Environment (MASSIVE) \texttt{(www.massive.org.au)}.
\newpage

\bibliographystyle{coling.bst}
\bibliography{coling2020}
\newpage

\section*{Appendix A:}
\section*{Experiments and Analysis}
\label{sec:appendix}
\subsection*{Data Preprocessing}
\label{sec:app-data}
We use the European languages of IWSLT\footnote{https://wit3.fbk.eu/} and TED\footnote{https://github.com/neulab/word-embeddings-for-nmt} datasets in our experiments as listed in Table \ref{tab:data-code}.
We filter the parallel corpus with \texttt{langid.py} \cite{lui2012langid} and remove sentences with a length ratio greater than 1.5.  All the sentences are first tokenized with the Moses tokenizer\footnote{https://github.com/moses-smt/mosesdecoder/blob/master/scripts/tokenizer/tokenizer.perl} and then segmented with BPE segmentation \cite{sennrich2015neural} with a learned BPE model by 32k merge operations on all
languages. We keep the output vocabulary of the teacher and student models the same to make the knowledge distillation feasible.
%%%%%%%%%%%%%%%%%

% \begin{table}[ht!]
%     \centering
%     \captionsetup{font=small}
%     \scalebox{0.56}{
%     \begin{tabular}{cccccccccccc}
%      \toprule
%         Language name & Russian & German &Italian & Spanish & Polish & Dutch& Basque&Galician&Belarusian	&Norwegian&Slovenian	\\
%      \midrule
%       Code &  ru & de &it& es& pl&nl & eu&gl&be&nb&sl\\
%      \midrule
%      size (\#sent(k)) & 153\textbackslash6.9\textbackslash5.5&
%      160\textbackslash7.2\textbackslash6.7& 
%      167\textbackslash7.5\textbackslash5.5&
%      169\textbackslash7.6\textbackslash5.5&
%      128\textbackslash5.8\textbackslash5.4&
%      153\textbackslash6.9\textbackslash5.3&
%      3.3\textbackslash0.3\textbackslash0.3&
%      8.4\textbackslash0.6\textbackslash1&
%      3.3\textbackslash0.2\textbackslash0.6&
%      14\textbackslash0.8\textbackslash0.8&
%      14.5\textbackslash1.4\textbackslash0.6\\
%       \bottomrule
%     \end{tabular}
%     }
%     \caption{Language names and statistics for bilingual resources (Language$\rightarrow$English), (train\textbackslash dev\textbackslash test)}
%     \label{tab:data-code}
%     \vspace{-8pt}
% \end{table}

\begin{table}[ht!]
    \centering
    \captionsetup{font=small}
    \scalebox{0.7}{
    \begin{tabular}{ccccccccc}
    \toprule
    \multicolumn{7}{c}{\textbf{High-resource Languages}}\\& \\
     \midrule
        Language name & Russian & German &Italian & Spanish & Polish & Dutch	\\
     \midrule
       Code &  ru & de &it& es& pl&nl\\
     \midrule
     size (\#sent(k)) & 153\textbackslash6.9\textbackslash5.5&
     160\textbackslash7.2\textbackslash6.7& 
     167\textbackslash7.5\textbackslash5.5&
     169\textbackslash7.6\textbackslash5.5&
     128\textbackslash5.8\textbackslash5.4&
     153\textbackslash6.9\textbackslash5.3&\\
      \toprule
      \multicolumn{7}{c}{\textbf{Low-resource Languages}}\\& \\
      \midrule
        Language name & Basque & Galician & Norwegian & Slovenian& Estonian	\\
      \midrule
        Code & eu&gl&nb&sl&et\\
     \midrule
     size (\#sent(k)) &
     3.3\textbackslash0.3\textbackslash0.3&
     8.4\textbackslash0.6\textbackslash1&
     14\textbackslash0.8\textbackslash0.8&
     14.5\textbackslash1.4\textbackslash0.6&
     7.7\textbackslash0.7\textbackslash1\\
      \bottomrule
    \end{tabular}
    }
    \caption{Language names and statistics for bilingual resources (Language$\rightarrow$English), (train\textbackslash dev\textbackslash test)}
    \label{tab:data-code}
    \vspace{-8pt}
\end{table}

%%%%%%%%%%%%%%%%%

% \subsection{Model Configuration}
% \label{sec:app-modelconfig}
% All models are trained with Transformer architecture \cite{vaswani2017attention}, with the model hidden size of 256, feed-forward hidden size of 1024, and 2 layers,  implemented in Fairseq framework \cite{ott2019fairseq}. 
% We use the Adam optimizer \cite{kingma2014adam}  and an inverse square root schedule with warm-up (maximum LR 0.0005). We apply dropout and label smoothing with a rate of 0.3 and 0.1. The source and target embeddings are shared and tied with the last layer. We train with half-precision floats on one V100 GPU, with at most 4028 tokens per batch. 

% \subsection{Contribution Weight Analysis}
% \label{sec:app-contrib-weight-analysis}
% In order to make clear how different teachers contribute to train the student (low-resource NMT model), we illustrate the contribution weights of all teachers for the first 30 iterations of different mini-batches while training the low-resource NMT model (Figure \ref{fig:contrib-wights}). The contribution weight plots show that for ``sl-en" and ``nb-en" pairs, teachers' contribution weights are very close together while for ``gl-en" we have one dominant teacher. As it is also shown earlier in the results, for those cases where most of the teachers are linguistically close, our approach outperforms other baselines by adaptively assigning the contribution weights to the teachers during the knowledge distillation process.
% \input{fig2-weights}
\subsection*{Translation Examples}
\label{sec:app-trans-examples}
Table \ref{tab:example-data-compressed} showcases the generated English translations by the individual student, all the teachers, and student trained through adaptive knowledge distillation from Norwegian language. This example shows that while there is a diversity between different teachers' translations e.g., for the verb of \emph{``provoke"}, the student is impacted by the agreement of the majority of teachers. Moreover, this example shows that our adaptive KD model captures the best of all teachers resulting in a higher quality translation.

\begin{table*}[ht!]
\centering
\small
\captionsetup{font=small}
  \scalebox{0.9}{
\begin{tabular}{|c|p{13cm}|}
\toprule
Ref & And great creativity is needed to do what it does so well : to provoke us to think differently with dramatic creative statements .\\ &\\
\midrule
Individual & \textcolor{red}{kepler} \textcolor{green}{\textbf{great}} \textcolor{red}{mission mission} to do it as \textcolor{green}{\textbf{well}} : to \textcolor{red}{grow} us  \textcolor{green}{to think \textbf{with dramatic}} creativity . \\
\midrule
Teacher (ru-en) &  and \textcolor{red}{the first} \textcolor{green}{\textbf{creativity} needed \textbf{to do what it does}} : to \textcolor{red}{promote} us to think about the \textcolor{green}{\textbf{dramatic}} creativity .
\\
\midrule
Teacher (de-en) & \textcolor{red}{now , the future} \textcolor{green}{\textbf{creativity}} needs to do it as it does : to \textcolor{green}{\textbf{provoke} us to \textbf{think differently} \textbf{with dramatic} creative} \textcolor{red}{expression} .
\\
\midrule
Teacher (it-en) & \textcolor{red}{ now , the future} \textcolor{green}{\textbf{creativity} is needed \textbf{to do what it does} so \textbf{well}} : to \textcolor{red}{provocate} \textcolor{green}{us to \textbf{think differently}} about \textcolor{green}{\textbf{dramatic}} \textcolor{red}{reactive} .
\\
\midrule
Teacher (es-en) & \textcolor{red}{the future of} \textcolor{green}{\textbf{creativity}} to \textcolor{red}{do that as it's doing so good} : to \textcolor{red}{provocate} us to \textcolor{green}{\textbf{think differently}} about \textcolor{green}{\textbf{dramatic}} creativity .

\\
\midrule
Teacher (pl-en) & \textcolor{red}{the future of} \textcolor{green}{\textbf{creativity}} \textcolor{green}{\textbf{to do what it does} so} \textcolor{red}{good} : to \textcolor{red}{promise} others \textcolor{green}{\textbf{with dramatic}} creativity .
\\
\midrule
Teacher (nl-en) & \textcolor{red}{now , the frequent} \textcolor{green}{\textbf{creativity}} is \textcolor{red}{to make it that it makes} so \textcolor{red}{good} : \textcolor{green}{to \textbf{provoke} us} with \textcolor{green}{\textbf{dramatic} creative} .
\\
\midrule
Proposed Adapt. KD & \textcolor{red}{now , they} need \textcolor{green}{\textbf{great} \textbf{creativity} \textbf{to do what it does} so \textbf{well}} : \textcolor{green}{\textbf{provoke} us to \textbf{think differently} \textbf{with dramatic}} creativity.
\\
\bottomrule
\end{tabular}}
\vspace{-.1cm}
\caption{The generated outputs from the individual student, all teachers, and student trained with multi-teachers (Proposed Adapt. KD) for ``nb-en" MT task. Some of the correct keyword translations are indicated with green color while hallucinations are represented by red. The bold-green shows the best of the teachers' output which is also captured with the student.}
\label{tab:example-data-compressed}
\end{table*}

\newpage

\end{document}